\documentclass[11pt,a4paper]{article}
\usepackage[hyperref]{acl2020}
\usepackage{times}
\usepackage{latexsym}

\usepackage{amsfonts}
\usepackage{amsmath}
\usepackage{booktabs}
\usepackage{multirow}
\usepackage{graphicx}
\usepackage{enumitem}
\usepackage{textcomp}
\usepackage{booktabs}

\usepackage{microtype}

\aclfinalcopy 
\def\aclpaperid{71} 

\newcommand*{\affaddr}[1]{#1}
\newcommand*{\affmark}[1][*]{\textsuperscript{#1}}
\newcommand*{\email}[1]{#1}

\title{You Impress Me: Dialogue Generation via Mutual Persona Perception}

\author{Qian Liu\affmark[\textdagger]{\thanks{~~Work done during an internship at Microsoft Research.}}~~, Yihong Chen\affmark[$\lozenge$]$^*$, Bei Chen\affmark[\S], Jian-Guang Lou\affmark[\S],\\ \textbf{Zixuan Chen}\affmark[$\spadesuit$]$^*$, \textbf{Bin Zhou\affmark[\textdagger], Dongmei Zhang\affmark[\S]}\\
\affaddr{\affmark[\textdagger]School of Computer Science and Engineering, Beihang University, China}\\
\affaddr{\affmark[$\lozenge$]UCL Centre for Artificial Intelligence, University College London, United Kindom}\\
\affaddr{\affmark[$\spadesuit$]School of Computer Science, Fudan University, China}\\
\affaddr{\affmark[\S]Microsoft Research, Beijing, China}\\
\affmark[\textdagger]\email{\{qian.liu, zhoubin\}@buaa.edu.cn; \affmark[\S]\{beichen, jlou, dongmeiz\}@microsoft.com};\\
\affmark[$\lozenge$]yihong.chen@cs.ucl.ac.uk;\,\affmark[$\spadesuit$]remch183@outlook.com
}

\date{}

\begin{document}
\maketitle
\begin{abstract}

Despite the continuing efforts to improve the engagingness and consistency of chit-chat dialogue systems, the majority of current work simply focus on mimicking human-like responses, leaving understudied the aspects of modeling understanding between interlocutors. The research in cognitive science, instead, suggests that understanding is an essential signal for a high-quality chit-chat conversation. Motivated by this, we propose $\mathcal{P}^2~\textsc{Bot}$, a transmitter-receiver based framework with the aim of explicitly modeling understanding. Specifically, $\mathcal{P}^2~\textsc{Bot}$ incorporates mutual persona perception to enhance the quality of personalized dialogue generation. Experiments on a large public dataset, \textsc{Persona-Chat}, demonstrate the effectiveness of our approach, with a considerable boost over the state-of-the-art baselines across both automatic metrics and human evaluations.

\end{abstract}

\section{Introduction}
\label{sec:intro}

Thanks to the advance in neural models and the accessibility of massive datasets, open-domain dialogue (i.e. chit-chat) systems have made great progress towards mimicking human-like responses. Nevertheless, there still exist some serious challenges in building personalized chatbots that can deliver engaging conversations and gain user trust \cite{song2019diverse}. For example, current chit-chat systems tend to generate uninformative responses \cite{li2016deep}. Moreover, they are usually lack of coherent personality traits due to the fact that training dialogues actually come from a diverse set of speakers \cite{zhang2018personalizing}. 

\begin{figure}
    \centering
    \includegraphics[width=.4\textwidth]{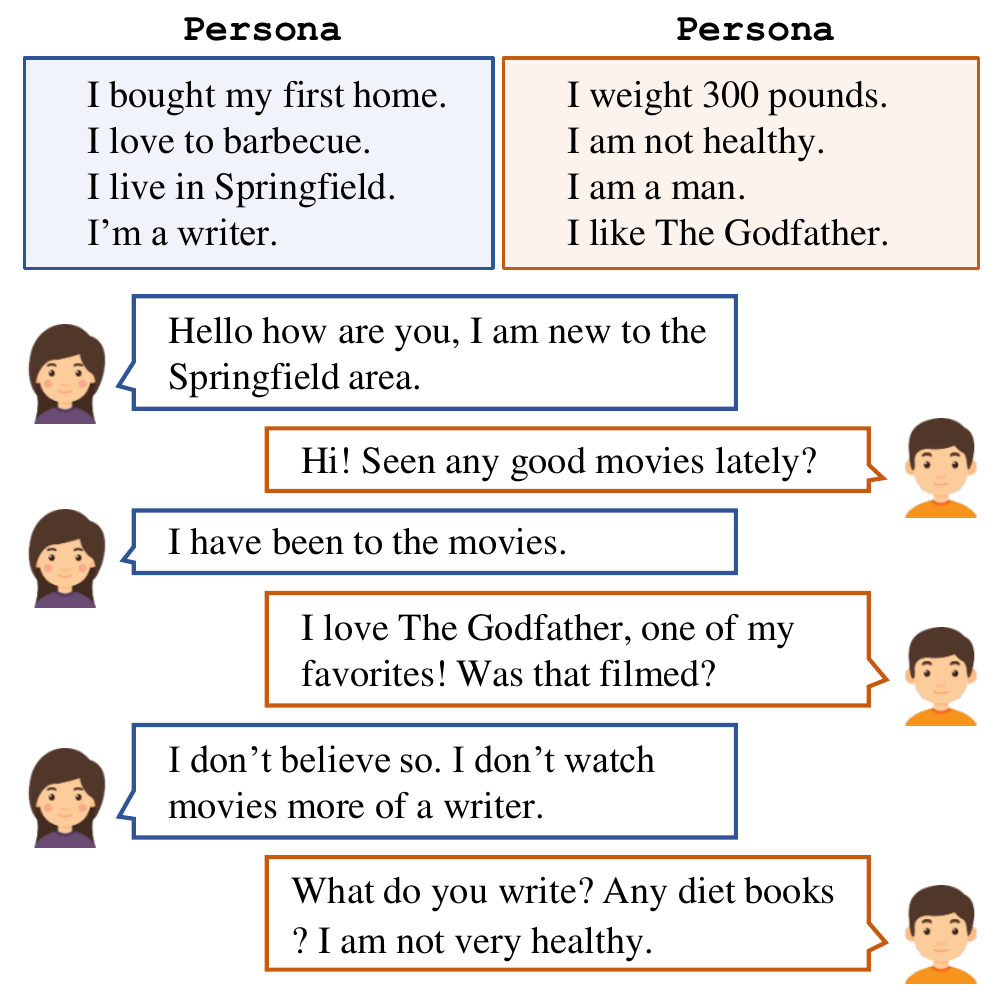}
    \caption{A clippled dialogue from \textsc{Persona-Chat}.}
    \label{fig:dialogue_example}
\end{figure}

Several attempts have been made to alleviate the above issues. Methods like special reward shaping to reduce generic responses \cite{li2016deep} and representing the speakers with latent variables \cite{li2016persona} were introduced to improve the engagingness of chit-chat systems. A more straightforward approach, which equips chit-chat systems with predefined personas, was proposed accompanied by a novel dataset, \textsc{Persona-Chat} \cite{zhang2018personalizing}. Figure \ref{fig:dialogue_example} shows a clipped dialogue from \textsc{Persona-Chat}. Two interlocutors meet for the first time and are having a conversation in order to get to know each other. What makes \textsc{Persona-Chat} unique is that personas of both interlocutors are explicitly described using several profile sentences, facilitating the training of chatbots with configurable and persistent personalities.

\begin{figure*}
    \centering
    \includegraphics[width=.8\textwidth]{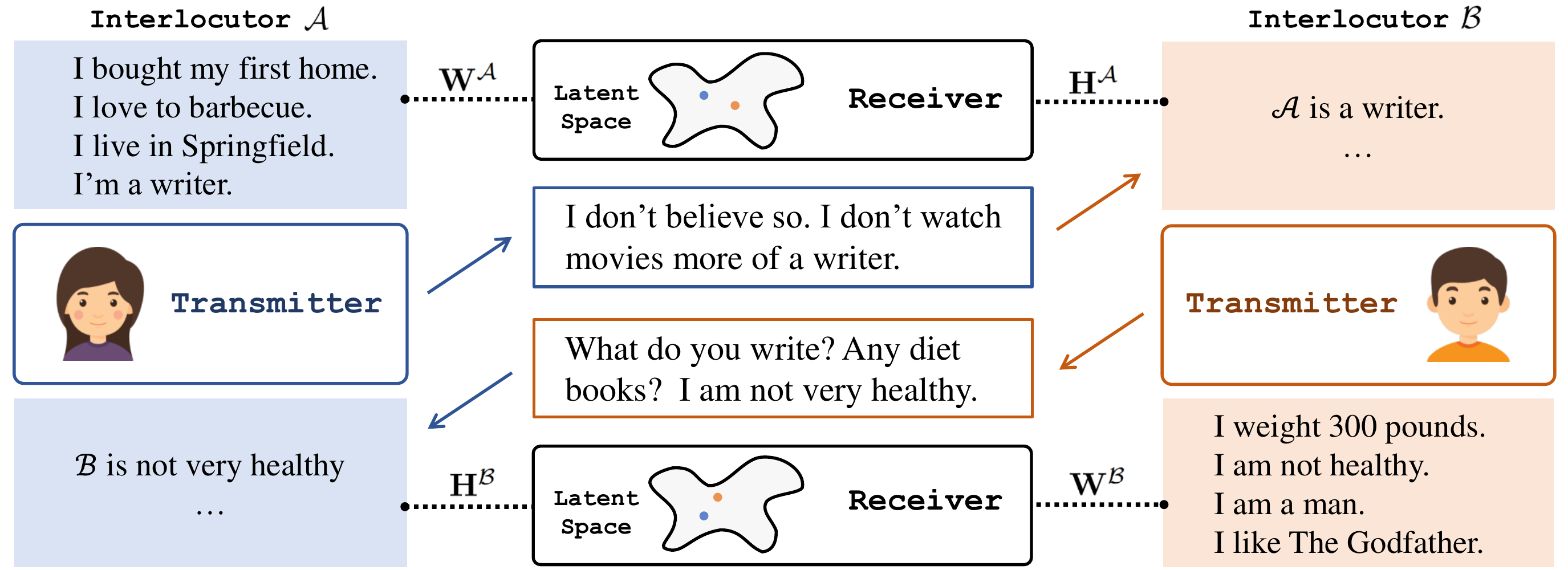}
    \caption{The overview of $\mathcal{P}^2~\textsc{Bot}$ (see text).}
    \label{fig:model_framework}
\end{figure*}

\textsc{Persona-Chat} has fueled a growing interest in developing methods for personalized dialogue generation. \citet{mazare2018training} incorporated additional data from Reddit to train the model. \citet{wolf2019transfertransfo} fine-tuned pretrained language model \cite{Radford2018ImprovingLU} to improve the dialogue generation. Although both works demonstrate promising results, they focus more on mimicking the style of human-like responses, leaving understudied the aspects of explicitly modeling understanding between interlocutors. Our work, instead, takes the perspective of understanding modeling.

According to the research in cognitive science, effective communication creates similar activation maps in the brains of both interlocutors~\cite{hasson2012brain}, suggesting that understanding between interlocutors is an essential signal for a high-quality chit-chat conversation. 
For instance, in the conversation shown in Figure~\ref{fig:dialogue_example}, the two interlocutors foster understanding either by raising persona-related topics, \textit{``Seen any good movies lately?''}, or by revealing their own personas through answering questions, \textit{``I don't watch movies more of a writer.''}. The efforts to build understanding keep the conversation flowing.

Taking into account the above, we propose Persona Perception Bot ($\mathcal{P}^2~\textsc{Bot}$), explicitly modeling the understanding between interlocutors with a transmitter-receiver framework. Distinguished from traditional methods, $\mathcal{P}^2~\textsc{Bot}$ highlights a novel concept, \textbf{mutual persona perception}, which is better suited to describe the information exchange process that empowers the interlocutors to get to know each other. In order to train $\mathcal{P}^2~\textsc{Bot}$ for personalized dialogue generation, we employ supervised training and self-play fine-tuning piloted by reward signals characterizing mutual persona perception. Experiments on the \textsc{Persona-Chat} dataset demonstrate the superiority of our approach over the baselines in both automatic metrics and human evaluations\footnote{Our code is available at \url{https://github.com/SivilTaram/Persona-Dialogue-Generation}}.

\section{Methodology Overview}

The central idea of $\mathcal{P}^2~\textsc{Bot}$ is to explicitly model understanding between interlocutors and enhance dialogue generation via mutual persona perception. It comprises two components, \textbf{Transmitter} and \textbf{Receiver}, respectively responsible for dialogue generation and mutual persona perception. Figure~\ref{fig:model_framework} gives an overview of $\mathcal{P}^2~\textsc{Bot}$: interlocutor $\mathcal{A}$ has a persona $\mathbf{w}^{\mathcal{A}}$, described with $L$ profile sentences $\{w^{\mathcal{A}}_{1}, \cdots, w^{\mathcal{A}}_{L}\}$. When she first meets the other interlocutor $\mathcal{B}$, they are going to know each other through a $N$-turn dialogue $(x^{\mathcal{A}}_{1}, x^{\mathcal{B}}_{1}, \cdots, x^{\mathcal{A}}_{N}, x^{\mathcal{B}}_{N})$, where $x^{\mathcal{A}}_{n}$ denotes the utterance that $\mathcal{A}$ says in $n$-th turn and $N$ denotes the number of total turns. Given the entire dialogue history up to $n$-th turn $\mathbf{h}^{\mathcal{A}}_n = (x^{\mathcal{A}}_{1}, \cdots, x^{\mathcal{B}}_{n-1})$, \textbf{Transmitter} generates $x^\mathcal{A}_{n}$ according to the distribution $p(x^\mathcal{A}_{n}\,|\,\mathbf{w}^\mathcal{A},\mathbf{h}^{\mathcal{A}}_n)$, and transmits it to $\mathcal{B}$. The same process applies to $\mathcal{B}$, keeping the conversation flowing.

As the conversation goes on, impressions are gradually built via utterances. 
For example, when $\mathcal{A}$ says \textit{``I don't watch movies more of a writer.''}, the impression that \textit{``$\mathcal{A}$ is a writer.''} is left on $\mathcal{B}$'s mind. 
As mentioned above, a successful conversation helps interlocutors know each other, 
which means $\mathcal{B}$'s impression of $\mathcal{A}$ should correspond to $\mathcal{A}$'s persona and vice versa. \textbf{Receiver} aims to measure the proximity between the built impressions and the actual personas. Specifically, as demonstrated by the dashed black lines in Figure~\ref{fig:model_framework}, Receiver first projects impressions and personas into a latent space, and then measures the relevance between them based on the \emph{impression encoding} (e.g. $\mathbf{H}^\mathcal{A}$, $\mathcal{B}$'s impression on $\mathcal{A}$, projected from $\mathcal{A}$'s utterances $\mathbf{x}^\mathcal{A}$), and \emph{persona encoding} (e.g. $\mathbf{W}^\mathcal{A}$, projected from $\mathcal{A}$'s persona $\mathbf{w}^\mathcal{A}$)\footnote{We take $\mathcal{A}$ as an example, and all are similar to $\mathcal{B}$.}. The relevance scores serve as mutual persona perception rewards, and are further incorporated into the training of Transmitter. Details of the two components are presented in Section \ref{sec-transmitter} and \ref{sec-receiver}. 

\section{Transmitter}\label{sec-transmitter}

Following previous work \cite{li2016deep,zhang2018personalizing}, we treat dialogue generation as a sequence generation problem. Concretely, we employ the pretraining transformer language model introduced in \citet{Radford2018ImprovingLU} (i.e. GPT) to initialize Transmitter. The entire training procedure consists of two steps: (1) \textbf{Supervised Dialogue Generation}. We optimize Transmitter via maximum likelihood estimation (MLE) on the supervised dialogue generation task. (2) \textbf{Self-play Model Fine-tuning}. We simulate dialogues between two randomly paired interlocutors, encouraging Transmitter to learn a policy that maximizes reward signals via reinforcement learning (RL)~\cite{sutton2000policy}. The design of the reward function considers both language modeling and our proposed mutual persona perception.

\subsection{Supervised Dialogue Generation}\label{sec:supervised}

\begin{figure}[t]
    \centering
    \includegraphics[width=.47\textwidth]{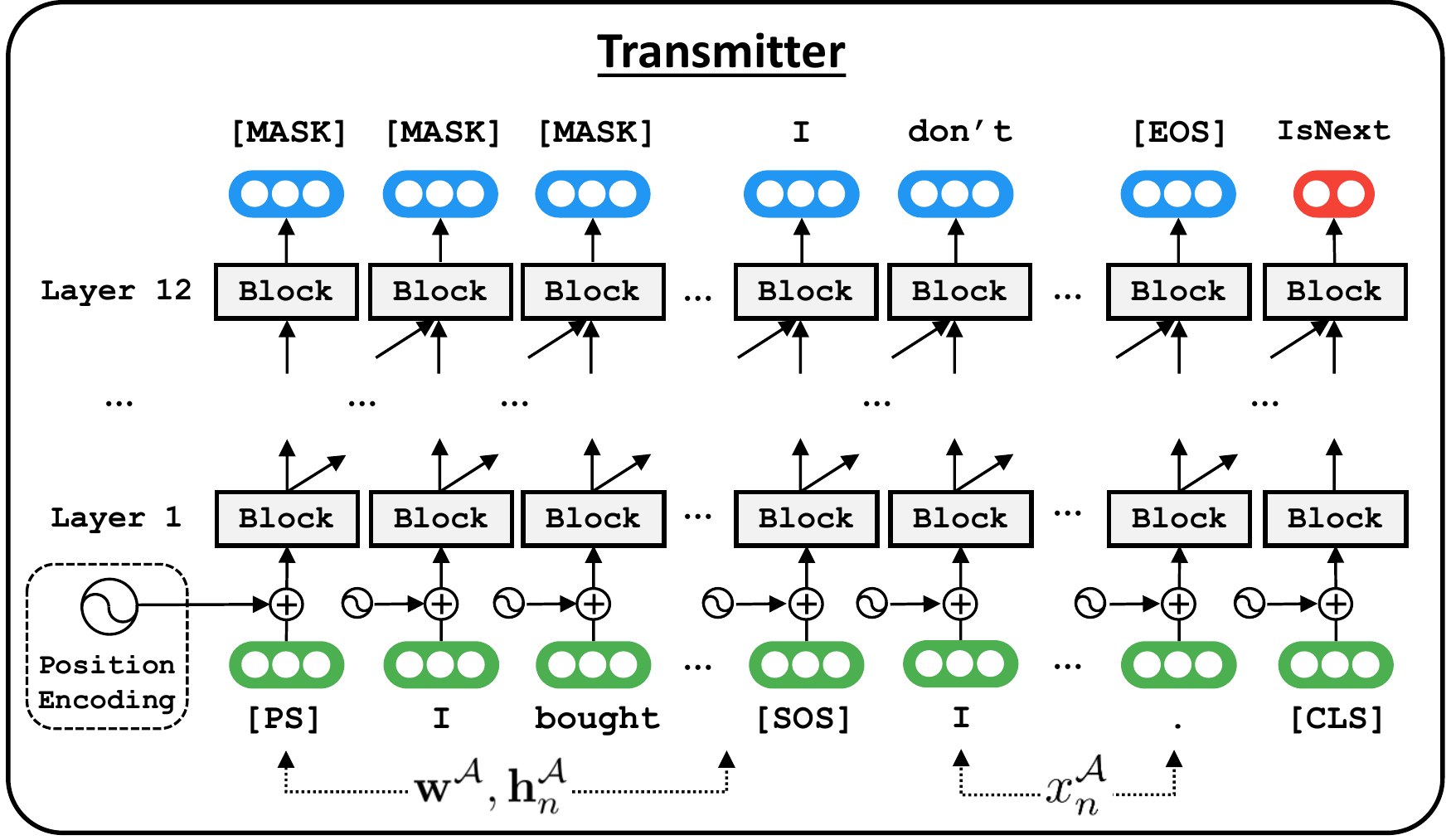}
    \caption{The overall architecture of Transmitter. ``Block'' is short for ``Transformer Block''. Arrows \scalebox{0.8}{$\nearrow$} bridge the current block to subsequent blocks of its following layer. Position encoding is to incorporate position information into block by assigning an embedding for each absolute position in the sequence. Here we omit the architecture inside the block, and refer the readers to \citet{vaswani2017attention} for more details. \texttt{[MASK]} tokens are ignored in the training objective.}
    \label{fig:transmitter_model}
\end{figure}

As illustrated in Figure~\ref{fig:transmitter_model}, Transmitter follows the overall architecture of 12 stacked transformer layers to encode context and generate response. Here, the context contains the persona $\mathbf{w}^{\mathcal{A}}$, the dialogue history $\mathbf{h}^{\mathcal{A}}_n$, and several special tokens (e.g. \texttt{[PS]} which indicates the start of persona). Given a training instance $(\mathbf{w}^{\mathcal{A}},\mathbf{h}^{\mathcal{A}}_n, x^{\mathcal{A}}_n)$, the training objective of MLE is to maximize the conditional log-likelihood as:
\begin{equation}\label{eq:mle_obj}
    \mathcal{L}_{\rm{mle}} = \sum_{t} \log p_{\theta}(x^\mathcal{A}_{n,t}\,|\,\mathbf{w}^{\mathcal{A}},\mathbf{h}^{\mathcal{A}}_n\,,x^\mathcal{A}_{n,<t}),
\end{equation}
where $\theta$ is the parameter of Transmitter. $x^\mathcal{A}_{n,t}$ means the $t$-th token in $x^\mathcal{A}_n$, and $x^\mathcal{A}_{n,<t}$ indicates the token sequence before $t$-th token. Equation~\ref{eq:mle_obj}, hereafter simplified as $\log p_{\theta}(x^\mathcal{A}_n\,|\,\mathbf{w}^{\mathcal{A}},\mathbf{h}^{\mathcal{A}}_n)$, applies to both $\mathcal{A}$ and $\mathcal{B}$, and we mention $\mathcal{A}$ for the sake of brevity (the same as below).

During inference, beam search is applied to store top-ranked response candidates $\{\hat{x}^{\mathcal{A}}_n\}$, and Transmitter subsequently chooses as prediction the one that maximizes the length-normalized score:
\begin{equation}\label{eq:max_response}
    x^{\mathcal{A}^*}_{n} = \mathop{\arg\max}_{\hat{x}^{\mathcal{A}}_{n}} \frac{\log p_{\theta}(\hat{x}^{\mathcal{A}}_n\,|\,\mathbf{w}^{\mathcal{A}},\mathbf{h}^{\mathcal{A}}_n)}{|\hat{x}^{\mathcal{A}}_n|}.
\end{equation}

Besides the sequence generation task, inspired by \citet{wolf2019transfertransfo}, we set up an auxiliary task, \textbf{Next Utterance Prediction}. Apart from training Transmitter to generate responses, we also train it to discriminate whether the response is the next utterance of the given context. Concretely, we append a special token \texttt{[CLS]} to the tail of the generated tokens. A classifier is built on top of the token's hidden state in the last transformer layer, as indicated by the red rounded rectangle in Figure \ref{fig:transmitter_model}. In training, for each response, we randomly sample a distractor and train the classifier to give a higher score on the response than the distractor. In inference, the classifier is used to rank response candidates together with Equation~\ref{eq:max_response}. Denoting as $y_n=1$ the signal indicating the generated response $\hat{x}^{\mathcal{A}}_n$ is predicted as the next utterance, Equation~\ref{eq:max_response} is extended as:
\begin{equation}\label{eq:final_score}
\begin{aligned}
     x^{\mathcal{A}^*}_{n}=\mathop{\arg\max}_{\hat{x}^{\mathcal{A}}_{n}}\bigg(\alpha&\!\cdot \frac{\log p_{\theta}(\hat{x}^{\mathcal{A}}_n\,|\,\mathbf{w}^{\mathcal{A}},\mathbf{h}^{\mathcal{A}}_n)}{|\hat{x}^{\mathcal{A}}_n|} \\
     + (1-\alpha)\cdot \log p_{{\theta}}(&{y_n}=1|\mathbf{w}^{\mathcal{A}},\mathbf{h}^{\mathcal{A}}_n,\hat{x}^{\mathcal{A}}_n)\bigg),
\end{aligned}
\end{equation}
where $\alpha$ is a hyper-parameter.

\subsection{Self-play Model Fine-tuning}

Although supervised dialogue generation alone can be used to mimic human-like responses, it does not inherently target at understanding. Therefore, we further fine-tune Transmitter using reinforcement learning with the goal of maximizing mutual persona perception. Analogous to \citet{lewis2017deal}, we apply \textbf{self-play} to simulate the communication between two Transmitters, both of which have been trained as described in Section \ref{sec:supervised}. 

Specifically, we have the two Transmitters communicate with each other for several turns. One Transmitter serves as a user with the parameters frozen, while the other is a learnable \textbf{agent}. The parameter of the learnable agent, $\theta$, is fine-tuned during the self-play. Without loss of generality, in our experiments, we let interlocutor $\mathcal{A}$, who starts a conversation, be the user, and correspondingly $\mathcal{B}$ be the learnable agent.

\begin{figure}
    \centering
    \includegraphics[width=.45\textwidth]{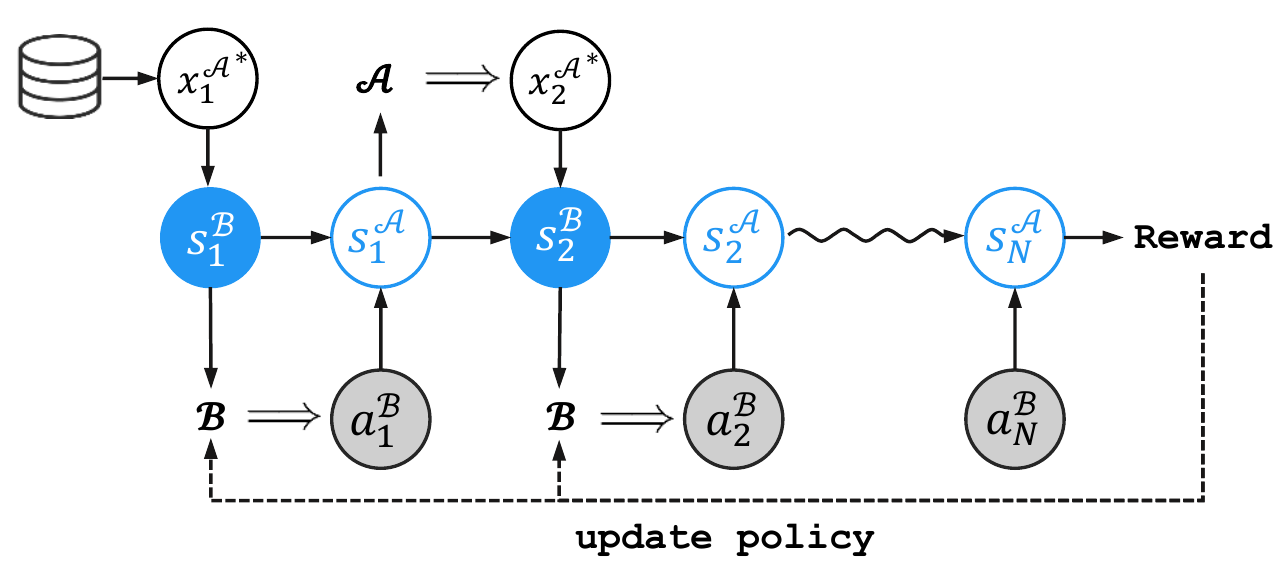}
    \caption{The illustration of the self-play procedure. Arrows $\Rightarrow$ represent the process of dialogue generation driven by Transmitter. Note that $x_1^{\mathcal{A}^*}$ is directly taken from the dataset as it is difficult to generate high-quality utterances without any dialogue history.}
    \label{fig:self_play}
\end{figure}

Here we introduce some necessary formulations for modeling our problem with reinforcement learning. 
A \textbf{state} contains the persona and the dialogue history. For example, the state for $\mathcal{B}$ at turn $n$ is defined as $s^{\mathcal{B}}_n=\{\mathbf{w}^{\mathcal{B}},\mathbf{h}^{\mathcal{B}}_n\}$. 
An \textbf{action} $a^\mathcal{B}_n$ is the response to be generated. The action space is infinitely large as the response can be arbitrary long. Taking $s^\mathcal{B}_n$ as input, the parameter $\theta$ defines a \textbf{policy} $p_{\theta}(a^\mathcal{B}_n|s^\mathcal{B}_n)$, through which the learnable agent generates its response.

As illustrated in Figure~\ref{fig:self_play}, when it is $\mathcal{B}$'s turn to speak, $\mathcal{B}$ receives $s^{\mathcal{B}}_n$ and picks $a^\mathcal{B}_n$ according to the policy $p_{\theta}$. As for $\mathcal{A}$, it receives $s^{\mathcal{A}}_n$ and generates the response $x^{\mathcal{A}^*}_{n}$ to simulate a user.
$\mathcal{A}$ and $\mathcal{B}$ alternately produce responses till the number of turns exceeds the given limit. Once a complete dialogue is generated, the reward is collected to optimize $\theta$ using policy gradient \cite{sutton2000policy}. Denoting as $R(a^\mathcal{B}_n)$ the reward $\mathcal{B}$ gets at turn $n$ (more details are provided later), we can optimize it by maximizing the following objective:
\begin{equation}\label{eq:l_rl}
    \mathcal{L}_{\rm{rl}} = \mathbb{E}_{a^\mathcal{B}_n{\sim}p_{\theta}(a^\mathcal{B}_n|s^\mathcal{B}_n)}[R(a^\mathcal{B}_n)].
\end{equation}
Applying likelihood ratio trick, $\theta$ is updated by ascending the following gradient:
\begin{equation}\label{eq:gradient_rl}
   \!{\nabla}_{\theta}{\mathcal{L}}_{\rm{rl}}\!=\!\mathbb{E}_{a^\mathcal{B}_n{\sim}p_{\theta}(a^\mathcal{B}_n|s^\mathcal{B}_n)}{\nabla}_{\theta}\!\log\!p_{\theta}(a^\mathcal{B}_n|s^\mathcal{B}_n)R(a^\mathcal{B}_n).\!\! \!\!\!\!
\end{equation}

As aforementioned, the space of action $a^\mathcal{B}_n$ is infinite. In practice, REINFORCE algorithm \cite{williams1992simple} is leveraged to approximate Equation~\ref{eq:gradient_rl} by sampling $a^\mathcal{B}_n$ from policy $p_{\theta}(a^\mathcal{B}_n|s^\mathcal{B}_n)$. Furthermore, subtracting a baseline \cite{weaver2001optimal}, here the mean reward of a mini-batch, is applied on $R(a^\mathcal{B}_n)$ to reduce variance. The agent samples tokens one by one through \emph{multinomial sampling} over the output distribution of $\mathcal{B}$, until the special token \texttt{[EOS]} is sampled or exceeding the maximum allowed decoding step (e.g. 32). Compared to beam search sampling, multinomial sampling provides more diversities.

\subsection{Reward Shaping (RS)}\label{sec:reward}

As described in Section \ref{sec:intro}, we believe that a high-quality chit-chat conversation should highlight both human language modeling and mutual persona perception. Bearing this in mind, we design three rewards to address language style, discourse coherence and mutual persona perception respectively.

\paragraph{RS.1 Language Style} The generated responses should conform to human language styles, which we believe can be evaluated by a pretrained language model (i.e. GPT). After length normalization, the score for $a^\mathcal{B}_n$ is given as:
\begin{equation}
    R_1(a^\mathcal{B}_n) = \frac{1}{|a^\mathcal{B}_n|}\sum\limits_{t}\log p_{\rm{lm}}(a^\mathcal{B}_{n,t}\,|\,a^\mathcal{B}_{n,<t}),
\end{equation}
where $a^\mathcal{B}_{n,t}$ and $a^\mathcal{B}_{n,<t}$ have similar denotation as the previously mentioned $x^\mathcal{A}_{n,t}$ and $x^\mathcal{A}_{n,<t}$.

\paragraph{RS.2 Discourse Coherence} The language score is evaluated individually, without considering the discourse coherence. However, a reasonable response should establish links in meaning with context, which is also an important aspect of human-like responses. To take into account the discourse coherence, we employ the well-trained Next Utterance Predictor (mentioned in Section~\ref{sec:supervised}). The reward is given by the log probability of $a^\mathcal{B}_{n}$ being the next utterance of $s^\mathcal{B}_{n}$:
\begin{equation}
    R_2(a^\mathcal{B}_n) = \log p_{\theta}(y_n=1\,|\,a^\mathcal{B}_{n},s^\mathcal{B}_{n}).
\end{equation}

\paragraph{RS.3 Mutual Persona Perception} 
RS.1 and RS.2 only steer the agent training process towards human-like responding. They do not explicitly encourage understanding between interlocutors. Therefore, we meticulously design the reward to characterize mutual persona perception. 
Contrast from RS.1 and RS.2, mutual persona perception is a long-term goal throughout the whole dialogue, 
meaning that the effect of current action might only play out some time later. 
For instance, receiving \textit{``what are your hobbies?''} from $\mathcal{B}$, it is highly likely that $\mathcal{A}$'s response is relevant to $\mathcal{A}$'s hobbies. 
This suggests that, not only $\mathcal{A}$'s response but also $\mathcal{B}$'s initial question contributes to mutual persona perception. 
Denoting as $\gamma$ the discount factor indicating how far ahead $\mathcal{B}$ looks, the reward of mutual persona perception for $a^\mathcal{B}_n$ is defined as:
\begin{equation}\label{eq:r_3}
    \begin{aligned}
        R_3(a^\mathcal{B}_n)\!=r(a^\mathcal{B}_n)+&\!\sum\limits_{k=n+1}^{N}\!\!\Big({\gamma}^{2(k-n)-1}r(x^\mathcal{A^*}_k)\\
        &+{\gamma}^{2(k-n)}r(a^\mathcal{B}_k~) \Big),
    \end{aligned}
\end{equation}
where $r(a^\mathcal{B}_n)$ is the persona perception score that $\mathcal{B}$ obtains in $n$-th turn, and $r(x^\mathcal{A^*}_k)$ is defined likewise. $r(a^\mathcal{B}_n)$ can be computed using a score function:
\begin{equation}\label{eq:r_3_score}
    r(a^\mathcal{B}_n) = \text{score}(a^\mathcal{B}_n, \mathbf{w}^{\mathcal{B}}).
\end{equation}
In $\mathcal{P}^2~\textsc{Bot}$, the score function comes from Receiver, which will be elaborated in Section \ref{sec-receiver}. The final reward $R(a^\mathcal{B}_n)$ for $a^\mathcal{B}_n$ is a weighted sum of the rewards listed above:
\begin{equation}
    R = \lambda_{1}R_1 + \lambda_{2}R_2 + \lambda_{3}R_3,
\end{equation}
where $\lambda_{1}$, $\lambda_{2}$ and $\lambda_{3}$ are hyper-parameters.

\section{Receiver}
\label{sec-receiver}

Receiver is devised to measure the proximity between the built impressions and the actual personas, implemented by negative sampling. Specifically, in training, we randomly sample a persona distractor $\mathbf{w}^{\mathcal{Z}}$. Receiver is trained to identify the real persona $\mathbf{w}^\mathcal{A}$ from $\{\mathbf{w}^\mathcal{A}, \mathbf{w}^\mathcal{Z}\}$. In inference, for each utterance, Receiver is responsible for providing a reasonable relevance score, to model our proposed mutual persona perception. The score subsequently joins the self-play fine-tuning on Transmitter as part of the rewards, as in Equation~\ref{eq:r_3}.

\begin{figure}[t]
    \centering
    \includegraphics[width=.47\textwidth]{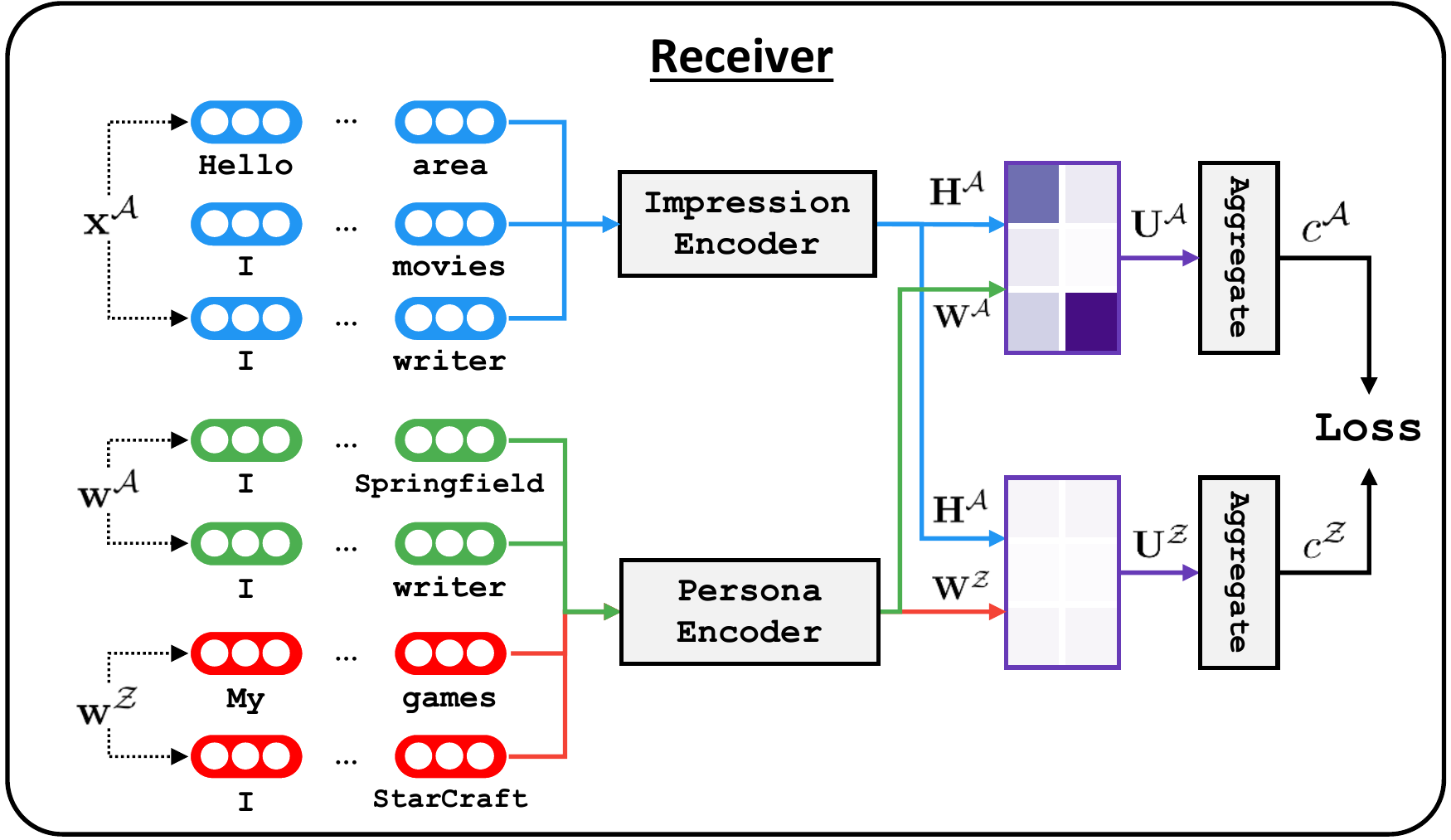}
    \caption{The\,overall\,architecture\,of\,Receiver\,(see\,text).}
    \label{fig:receiver_model}
\end{figure}

\subsection{Training}

As illustrated in Figure~\ref{fig:receiver_model}, Receiver contains two different encoders for impression and persona respectively. Initialized by BERT \cite{devlin2018bert}, both encoders provide deep contextualized representations for each token. Then we average all the representations, yielding a fixed $d$-dimensional vector for one sentence. In this way, feeding $(x^{\mathcal{A}}_{1},x^{\mathcal{A}}_{2},\cdots, x^{\mathcal{A}}_{N})$ into the impression encoder consecutively, we obtain the impression encoding $\mathbf{H}^{\mathcal{A}}\,{\in}\,\mathbb{R}^{N \times d}$. The persona encoding $\mathbf{W}^{\Delta}\,{\in}\,\mathbb{R}^{L \times d}$ is produced likewise, where $\Delta\in\!\{\mathcal{A},\mathcal{Z}\}\!$. The relevance score matrix $\mathbf{U}^{\Delta}$ is computed via the scaled dot product \cite{vaswani2017attention}:
\begin{equation}
    \mathbf{U}^{\Delta} = \frac{\mathbf{H}^\mathcal{A} ({\mathbf{W}^\Delta})^\top}{\sqrt{d}}, \,\in\,\mathbb{R}^{N{\times}L}.
\end{equation}

\begin{table*}[t]
    \centering
    \scalebox{0.82}{
    \begin{tabular}{llllllll}
        \toprule
        \multirow{2}{*}{\textbf{Category}} &
        \multicolumn{1}{c}{\multirow{2}{*}{\textbf{Model}}} & \multicolumn{3}{c}{\textbf{Original}} & \multicolumn{3}{c}{\textbf{Revised}} \\
     	\cmidrule(lr){3-5}
     	\cmidrule(lr){6-8}
         & & {\small Hits@1(\%)\,$\uparrow$} & {\small ~~ppl\,$\downarrow$} & {\small ~~F1(\%)\,$\uparrow$} & {\small Hits@1(\%)\,$\uparrow$} & {\small ~~ppl\,$\downarrow$} & {\small ~~F1(\%)\,$\uparrow$} \\
         \toprule
        \multirow{2}{1cm}{Retrieval} & KV Profile Memory & $54.8$ & ~~~~- & $14.25$ & $38.1$ & ~~~~- & $13.65$ \\
         & Dually Interactive Matching & $78.8$ & ~~~~- & ~~~~- & $\mathbf{70.7}$ & ~~~~- & ~~~~- \\ 
        \midrule
        \multirow{3}{1cm}{Generative} & Generative Profile Memory & $10.2$ & $35.01$ & $16.29$ & \,\,\,$9.9$ & $34.94$ & $15.71$ \\
        & Language Model & ~~~~- & $50.67$ & $16.30$ & ~~~~- & $51.61$ & $13.59$ \\
        & \textsc{Seq2seq-Attn} &  $12.5$ & $35.07$ & $16.82$ & \,\,\,$9.8$ & $39.54$ & $15.52$ \\
        \midrule
        \multirow{3}{1cm}{Pretrain\\Fintune} & Lost In Conversation & $17.3$ & ~~~~- & $17.79$ & $16.2$ & ~~~~- & $16.83$ \\
        & Transfertransfo & $\mathbf{82.1}$ & $17.51$ & $19.09$ & ~~~~- & ~~~~- & ~~~~- \\
        & $\mathcal{P}^2~\textsc{Bot}$ (Our) & $81.9$\,{\scriptsize [0.1]} & $\mathbf{15.12}$\,{\scriptsize [0.16]} & $\mathbf{19.77}$\,{\scriptsize [0.08]} & $68.6$\,{\scriptsize [0.2]} & $\mathbf{18.89}$\,{\scriptsize [0.11]} & $\mathbf{19.08}$\,{\scriptsize [0.07]} \\
        \bottomrule
        \end{tabular}
    }
    \caption{Automatic evaluation results of different methods on the \textsc{Persona-Chat} dataset. The standard deviation [$\sigma$] (across 5 runs) of $\mathcal{P}^2~\textsc{Bot}$ is also reported. All the results were evaluated on the dev set since the test set was not publicly available.}
    \label{tab:experiment_transmitter}
\end{table*}

In essence, Receiver is expected to capture fine-grained correlations between the persona and the dialogue. However, we do not have access to the golden fine-grained correlations. The only thing we know is that, compared with $\mathbf{W}^{\mathcal{Z}}$, $\mathbf{H}^{\mathcal{A}}$ is more correlated to $\mathbf{W}^{\mathcal{A}}$. Since the comparison is at a coarse granularity, we gather $\mathbf{U}^{\Delta}$ into the cumulative score $c^{\Delta}$ through an aggregate function $Agg$, as shown in Figure~\ref{fig:receiver_model}. To encourage $c^{\mathcal{A}}$ while at the same time depress $c^{\mathcal{Z}}$, we design a marginal loss $\mathcal{L}_{\rm{rec}}$, which makes $c^{\mathcal{A}}$ larger than $c^{\mathcal{Z}}$ by a margin $m$. Moreover, considering that an utterance generally relates to zero or one profile, $L_1$ regularization is enforced to make $\mathbf{U}^{\Delta}$ sparse. Combining all of these, the training loss for Receiver is:
\begin{equation}
    \mathcal{L}_{\rm{rec}} = \max(0, m + c^{\mathcal{Z}} - c^{\mathcal{A}}) + \beta \cdot |\mathbf{U}^{\Delta}|_{1},
\end{equation}
where $\beta$ is a hyper-parameter for penalty.

As for $Agg$, one straightforward way is to average over all positions of $\mathbf{U}^{\Delta}$. However, it maximizes every entry in $\mathbf{U}^{\mathcal{A}}$, including all those that should not be activated (e.g. relevance scores between unrelated profile sentences and utterances), introducing unnecessary noise into the training of Transmitter. To alleviate the problem, we choose to implement $Agg$ as a controllable weighted function, which summarizes $\mathbf{U}^{\Delta}_{n,:}$ as:
\begin{equation}
    Agg(\mathbf{U}^{\Delta}_{n,:}) = \frac{\sum\nolimits_{k=1}^{L}\exp(\mathbf{U}^{\Delta}_{n,k}/ \!\tau) \cdot \mathbf{U}^{\Delta}_{n,k}}{\sum_{k=1}^{L}\exp(\mathbf{U}^{\Delta}_{n,k}/ \!\tau)},
\end{equation}
where \emph{temperature} $\tau > 0$ is a tunable parameter \cite{hinton2015distilling} controlling the evolution of $Agg$. In the beginning, $Agg$ behaves close to average pooling. As $\tau$ anneals, $Agg$ gradually focuses more on the highest relevance score. In this way, noise reduces as training goes on. 
Finally, $c^{\Delta}$ is given by:
\begin{equation}
c^{\Delta} = \frac{1}{N}\sum\limits_{n=1}^{N}Agg(\mathbf{U}^{\Delta}_{n,:}).
\end{equation}

\subsection{Inference}

Given $x^\mathcal{A}_{n}$ and $\mathbf{w}^{\mathcal{A}}$, Receiver employs the following function to obtain $x^\mathcal{A}_{n}$'s persona perception score, further modeling mutual persona perception as in Equation \ref{eq:r_3_score}:
\begin{equation}
    \text{score}(x^\mathcal{A}_{n},\mathbf{w}^{\mathcal{A}})=\frac{Agg\big(\mathbf{H}^\mathcal{A}_{n,:} ({\mathbf{W}^\mathcal{A}})^\top\big)}{\sqrt{d}},
\end{equation}
where $\mathbf{H}^\mathcal{A}_{n,:}$ and ${\mathbf{W}^\mathcal{A}}$ are the impression encoding and persona encoding for $x^\mathcal{A}_{n}$ and $\mathbf{w}^{\mathcal{A}}$ respectively.

\section{Experiment}

We conducted experiments on the dataset \textsc{Persona-Chat}, assessing $\mathcal{P}^2~\textsc{Bot}$ using both automatic metrics and human evaluations. 
To verify the effectiveness of our proposed mutual persona perception, we perform a thorough model analysis in Section \ref{sec:model analysis}. Finally, we probe Receiver's capability on perceiving persona in Section \ref{sec:receiver discussion}.

\subsection{Implementation Details}

\textsc{Persona-Chat} dataset contains 8,939\,/\,1,000 multi-turn dialogues conditioned on 1,155\,/\,100 personas for train\,/\,dev. Each persona is described with at least 5 profile sentences. To make it more challenging, \textsc{Persona-Chat} also provides \textit{revised} personas by rephrasing, generalizing or specializing the \textit{original} ones. For example, \textit{``I am overweight.''} is revised from \textit{``I weight 300 pounds.''}.

Our implementation was based on PyTorch \cite{paszke2017automatic}, ParlAI \cite{miller2017parlai}, and HuggingFace's transformers library \cite{Wolf2019HuggingFacesTS}. We used Adam \cite{kingma2014adam} optimizer with a learning rate of 6.25e-5 for both Receiver and Transmitter in supervised learning. In the training of Receiver, $\tau$ reduced linearly from 10 to 0.5. In the self-play phase of Transmitter, the learning rate was set as 1e-6. The hyper-parameters $m$, $\alpha$, $\beta$, $\gamma$, $\lambda_1$, $\lambda_2$ and $\lambda_3$ were set as 0.4, 0.1, 1e-4, 0.5, 0.4, 0.1 and 0.5 respectively. The supervised training of Transmitter lasted for 2 epochs, and the self-play fine-tuning comprised 2000 dialogues, where the number of turns was 3. The beam search size was set as 2.

\subsection{Methods Comparison}\label{sec:model_compare}

Our baselines fall into three categories: retrieval-based, generative-based and pretrain-finetune-based models. Among the retrieval-based baselines, \textit{KV Profile Memory} \cite{zhang2018personalizing} was the official baseline which employed the memory network along with profile information, and \textit{Dually Interactive Matching Network} \cite{gu2019dually} proposed a dual matching architecture to match between the responses and their corresponding contexts. \textit{Language Model}, \textit{Generative Profile Memory} \cite{zhang2018personalizing} and \textsc{Seq2Seq} with attention mechanism \cite{Bahdanau2014NeuralMT} were implemented as generative baselines for dialogue generation. The remaining methods were all pretrain-finetune-based. \textit{Transfertransfo} \cite{wolf2019transfertransfo}\footnote{http://github.com/huggingface/transfer-learning-conv-ai} achieved the state-of-the-art performance on automatic metrics, while \textit{Lost In Conversation}\footnote{http://github.com/atselousov/transformer\_chatbot} topped the human evaluations \cite{dinan2019second}. Analogous to our approach, they employed the pretrained language model GPT to initialize their models, and then fine-tuned it on the dataset.

\begin{table}[t]
    \centering
    \scalebox{0.75}{
        \begin{tabular}{c|cccc|c}
            \toprule
            Model & 1\,(\%) & 2\,(\%) & 3\,(\%) & 4\,(\%) & Avg\\
            \midrule
            Lost In Conversation\!& $26.3$ & $\mathbf{48.7}$ & $22.0$ & ~~$3.0$ & $2.017$ \\
            Transfertransfo & $\mathbf{41.7}$ & $25.3$ & $\mathbf{28.7}$ & ~~$4.3$ & $1.956$ \\
            $\mathcal{P}^2~\textsc{Bot}$ (Our) & $18.9$ & $26.3$ & $28.6$ & $\mathbf{26.2}$ & $\mathbf{2.621}$ \\
            \bottomrule
        \end{tabular}
    }
    \caption{Human evaluation results.}
    \label{tab:human_evalution}
    \vspace{-2mm}
\end{table}

Table~\ref{tab:experiment_transmitter} shows the experimental results on automatic metrics. Following \citet{zhang2018personalizing}, we reported the official automatic metrics to evaluate the methods: \textbf{Hits@1}, \textbf{Perplexity (ppl)} and \textbf{F1}. Given 20 response candidates, Hits@1 is the probability that the real response ranks the highest according to the model. Perplexity measures the negative log likelihood of the correct sequence output by the model, lower values indicating better performance. F1 is the harmonic mean of word-level precision and recall. As observed, our approach outperforms almost all baselines and achieves new state-of-the-art performance on ppl and F1, with highly competitive performance on Hits@1. In the revised mode, our approach still achieves the best performance, obtaining a relative improvement of $13.4\%$ on F1 against the strongest baseline. It is worth noting that we also tried to employ F1 as the reward, but the result is far from satisfactory. 

As mentioned in \citet{dinan2019second}, no automatic metric is perfect for evaluating such an open-domain task. Hence, we also performed crowd-sourced \textbf{human evaluations} on the state-of-the-art baselines (i.e. Transfertransfo \& Lost In Conversation) and our proposed $\mathcal{P}^2~\textsc{Bot}$. Concretely, on the original dev set, we randomly sampled 200 responses generated by these methods and asked each worker to rate them. The rating ranges from $1$ to $4$. $1$ means the response is good only in terms of grammar and sentence structure; $2$ means in addition to valid grammar, the response is also coherent with the context; $3$ means the coherent response is meanwhile interesting and informative, instead of just a simple response like “Yes”; And $4$ means the response is consistent with the persona of the interlocutor, which is of extreme importance for the task of reflecting whether the model can effectively utilize the persona information. 
As shown in Table~\ref{tab:human_evalution}, the results are consistent with the automatic evaluation results, demonstrating the superiority of $\mathcal{P}^2~\textsc{Bot}$ against the baselines. We also conducted Wilcoxon signed-rank tests between our method and the baselines and the results show the improvements are significant with p $<$ $0.05$.

\begin{table}[t]
    \centering
    \scalebox{0.8}{
        \begin{tabular}{l|cll}
            \toprule
            Variant & {\small Hits@1(\%)\,$\uparrow$} & {\small F1(\%)\,$\uparrow$} & {\small BLEU(\%)\,$\uparrow$}\\
            \midrule
            $\mathcal{P}^2~\textsc{Bot}$-S & $68.7$ & $18.14$ & $0.56$ \\
            \midrule
            \;\!-\,\,Persona & $65.5$ & $17.77$\small{~(-\,$2.0$\%)} & $0.57$\small{~(+\,\,\,$1.8$\%)} \\
            \;\!-\,\,Next & $17.6$ & $18.11$\small{~(-\,$0.1$\%)} & $0.55$\small{~(-\,\,\,\,$1.8$\%)} \\
            + RS.1 & $68.4$ & $18.32$\small{~(+$0.9$\%)} & $0.60$\small{~(+\,\,\,$7.1$\%)} \\
            $\hookrightarrow$ + RS.2 & $68.6$ & $18.41$\small{~(+$1.5$\%)} & $0.61$\small{~(+\,\,\,$8.9$\%)} \\
            ~~~~~~$\hookrightarrow$ + RS.3 & $68.6$ & $19.08$\small{~(+$5.2$\%)} & $0.75$\small{~(+$33.9$\%)} \\
            \bottomrule
        \end{tabular}
    }
    \caption{Variant analysis results on \textsc{Persona-Chat} revised mode, along with relative improvements (shown inside brackets) compared with $\mathcal{P}^2~\textsc{Bot}$-S. BLEU refers to the cumulative 4-gram\! BLEU\! score. ``-\,Persona'' means dialogue generation without personas; ``-\,Next'' ablates the auxiliary task mentioned in Section~\ref{sec:supervised}; ``+\,RS.1'' means only using Language Style score as the reward in the self-play fine-tuning phase; ``$\hookrightarrow$ +\,RS.2'' means adding Discourse Coherence to the reward on the basis of RS.1; ``$\hookrightarrow$ +\,RS.3'' is equivalent to our proposed $\mathcal{P}^2~\textsc{Bot}$. }
    \label{tab:variant_analysis_transmitter}
\end{table}

\subsection{Model Analysis}\label{sec:model analysis}

\begin{table*}[t]
    \centering
    \setlength\tabcolsep{1pt}
    \begin{tabular}{c}
    \includegraphics[width=0.98\textwidth]{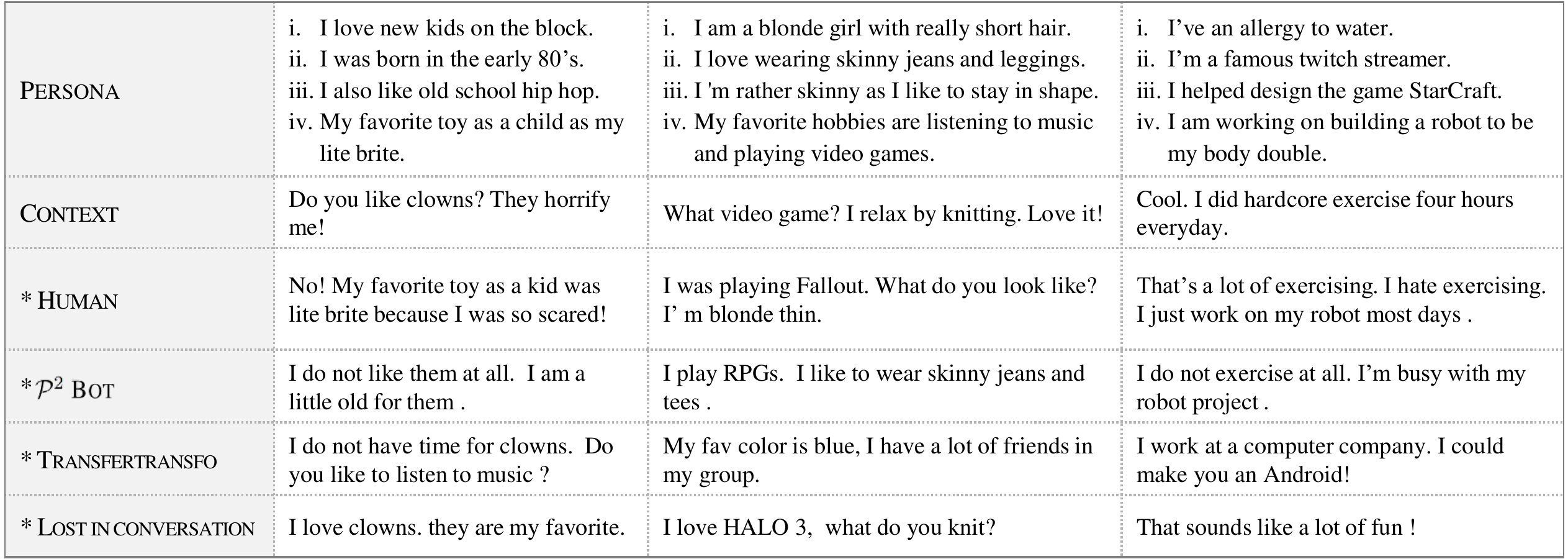}
    \end{tabular}
    \caption{Sampled responses(*) by Human, $\mathcal{P}^2~\textsc{Bot}$ and the state-of-the-art baselines.}
    \label{tab:generated_dialogue}
\end{table*}

\paragraph{Variant Analysis} We conducted variant analysis on $\mathcal{P}^2~\textsc{Bot}$ to investigate the influence of RS.1, RS.2 and RS.3. Another metric \textbf{BLEU} \cite{papineni-etal-2002-bleu}, which evaluates the quality of response, was introduced to make the analysis more comprehensive. We show the variant analysis results in Table~\ref{tab:variant_analysis_transmitter}, where $\mathcal{P}^2~\textsc{Bot}$-S is the variant of $\mathcal{P}^2~\textsc{Bot}$ which is trained only in the supervised setting. As expected, the results on Hits@1 validate the important role of the auxiliary task. Across all the variants, the gains in BLEU and F1 are very small, revealing the difficulty in improving them. Nevertheless, solely by adding RS.3, we obtained a $25\%$ relative improvement on BLEU, indicating the effectiveness of our proposed mutual persona perception. Similar conclusions can be drawn from the trend of F1.

\paragraph{Case Study} For a more comprehensive comparison, we show in Table~\ref{tab:generated_dialogue} some randomly sampled responses of different methods. The results suggest the responses generated by our approach are more human-like. As observed, benefiting from our proposed mutual persona perception, the responses of $\mathcal{P}^2~\textsc{Bot}$ are more consistent, engaging and informative. For instance, in the last example in Table~\ref{tab:generated_dialogue}, the response \textit{``I'm busy with my robot project''} explicates why the speaker does not exercise, meanwhile revealing that he is working on the robot, as depicted in his persona.

\paragraph{Error Analysis} Though our approach works well in most cases, we observed that the self-play simulation might fall into repeated cycles after rounds of training, as the challenge mentioned by~\citet{li2016deep}. Another issue is that the bots sometimes ask redundant questions in our approach, which might be due to inappropriate hyper-parameters in reward shaping.

\subsection{Persona Perception Probing}\label{sec:receiver discussion}

\begin{table}[t]
    \centering
    \scalebox{0.8}{
    \begin{tabular}{lcccc}
    \toprule
     \multicolumn{1}{c}{\textbf{Model}} & \multicolumn{2}{c}{\textbf{Original}} & \multicolumn{2}{c}{\textbf{Revised}} \\
 	\cmidrule(lr){2-3}
 	\cmidrule(lr){4-5}
      & {\small Hits@1\,$\uparrow$} & {\small MRR\,$\uparrow$} & {\small Hits@1\,$\uparrow$} & {\small MRR\,$\uparrow$} \\
    \midrule
Random & ~~$3.1$ & ~~$0.2$ & ~~$3.1$ & ~~$0.2$ \\
    IR & $67.5$ & $20.9$ & ~~$9.7$ & ~~$2.2$ \\
    \midrule
    Receiver & \textbf{93.8} & \textbf{37.5} & \textbf{78.2} & \textbf{16.6} \\
    \bottomrule
    \end{tabular}
    }
    \caption{Experimental results on Persona Perception.}
    \label{tab:receiver_exper}
\end{table}

\begin{figure}[t]
    \centering
    \includegraphics[width=.45\textwidth]{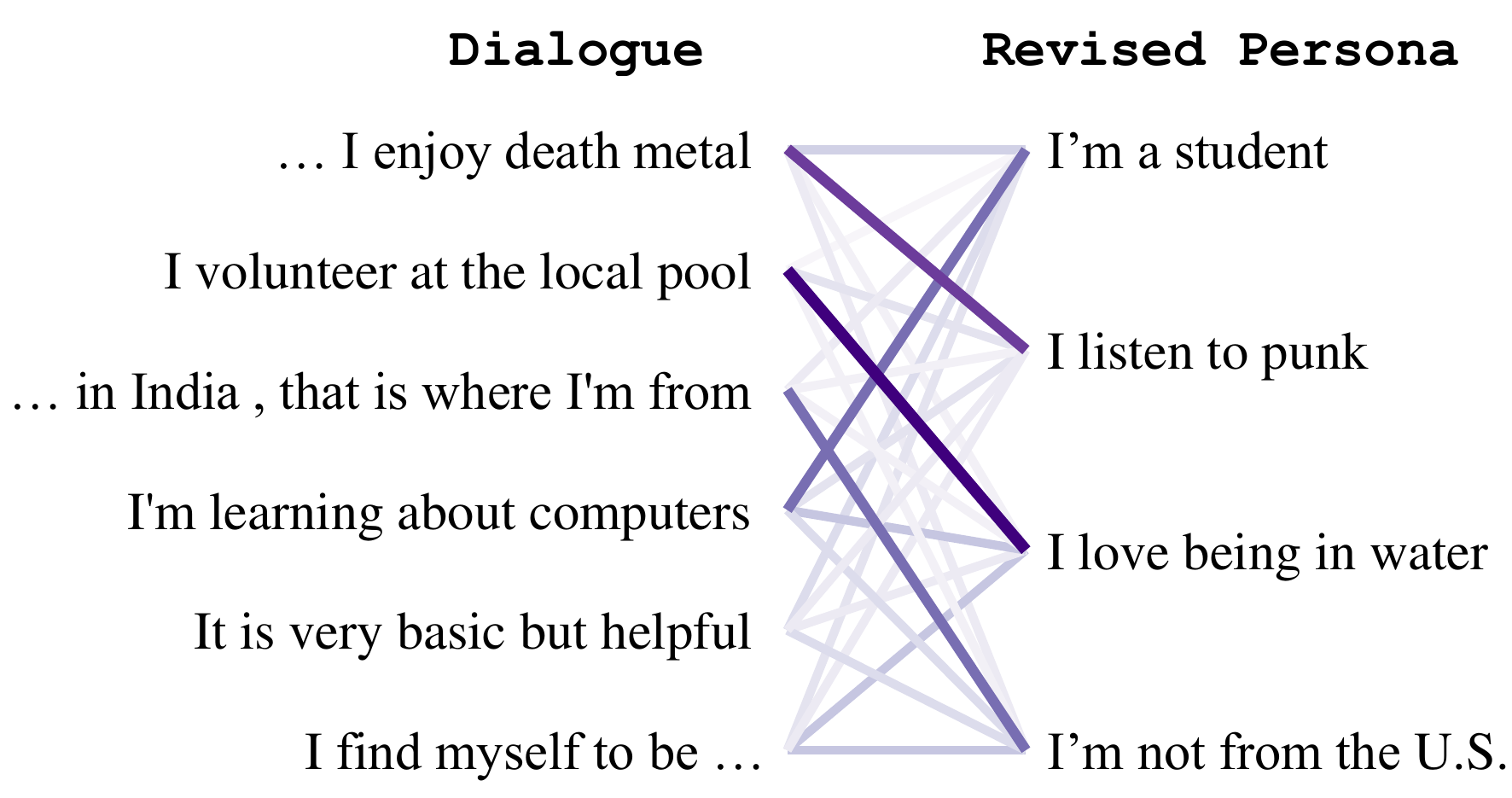}
    \caption{Visualization of the relevance scores between a sampled dialogue and its corresponding revised persona. Deeper color means higher score. We omit some context due to space limitation.}
    \label{fig:receiver_heatmap}
\end{figure}

Receiver plays an important role in our approach, and we are interested in its capability on perceiving persona. Therefore, we conducted experiments on a synthesized dataset. 
We constructed the dataset by sampling 31 persona distractors for each dialogue in \textsc{Persona-Chat}. Two widely used ranking metrics were used to evaluate the performance: \textbf{Hits@1} and \textbf{Mean Reciprocal Rank (MRR)}. Hits@1 is the same metric as the one mentioned in Section~\ref{sec:model_compare}, except that the candidate size is 32. 
Given a dialogue and the complete set of profile sentences, MRR is the average reciprocal ranks of the dialogue-relevant profile sentences. Two simple baselines Random and IR \cite{sordonineural2015} were chosen for comparison. Table~\ref{tab:receiver_exper} shows the experimental results of different methods on the synthesized dataset. As observed, our approach achieved excellent results on both original and revised modes. For example, compared with the IR baseline, our approach achieved an absolute improvement of $26.3\%$ on Hits@1 in the original mode. In addition, the surprising results in the revised mode further demonstrate Receiver's capability to perceive rephrased persona.

To further understand the trained Receiver, we visualize the relevance scores between a sampled dialogue and its corresponding revised persona in Figure~\ref{fig:receiver_heatmap}. As illustrated, the relevance scores between related profile sentences and dialogue utterances are significantly higher. For example, the utterance \textit{``I volunteer at the local pool''} from the interlocutor implies the profile \textit{``I love being in the water''}, and our Receiver successfully captures the relevance between them.

\section{Related Work}

Methods to build open-domain dialogue systems generally fall into two major categories: retrieval-based and generative-based. Retrieval-based methods retrieve response candidates and rank them based on the matching scores with the dialogue \cite{sordonineural2015,wu2017sequential,gu2019dually}. Generative-based methods typically use \textsc{Seq2Seq} model as the backbone \cite{sutskever2014sequence,Bahdanau2014NeuralMT,serban2017hierarchical,wolf2019transfertransfo},
where the encoder extracts the information in an utterance and the decoder generates the response. Our work adopts a similar architecture. Besides supervised learning, researchers also explore reinforcement learning based methods. \citet{lewis2017deal} applied reinforcement learning for negotiation dialogues and showed it outperforms supervised learning when negotiating with humans. \citet{yang2018personalized} proposed to generate dialogue responses by dual learning based domain adaptation. \citet{zhang2018reinforcing} built a coherence model to provide the reward signal for penalizing dull responses. \citet{liu-etal-2019-split} employed reinfrocement learning to learn an intermediate structure span. Our approach differs from this line of work in that we focus on improving personalized dialogues via mutual persona perception, which has not yet been explored before. 

More recently, under the topic of dialogue personalizing, \citet{zemlyanskiy2018aiming} proposed a post-processing method to re-rank candidates generated by beam search, while \citet{moradversarial} employed adversarial approaches to solve the consistency problem on interlocutors' names. \citet{madotto2019personalizing} applied meta-learning to quickly adapt to new speakers, and \citet{tigunova2019listening} extracted user attributes from daily dialogues. Compared with them, our work enhances persona based dialogue generation from a novel perspective. 

Furthermore, researchers explored to generate diverse responses conditioned on persona \cite{song2019diverse,song-aaai2020-generating}. Personalization in goal-oriented dialogue systems has also received some attention \cite{joshi2017personalization,luo2018learning}. The researches focus more on making the goal-oriented bots adjust the response according to different user profiles, while we aim to endow bots with persistent personalities.

\section{Conclusion \& Future Work}

We propose $\mathcal{P}^2~\textsc{Bot}$, a transmitter-receiver framework which explicitly models understanding between interlocutors. Under this framework, mutual persona perception is incorporated as a reward signal to achieve the personalized dialogue generation. Experiments on a large public dataset \textsc{Persona-Chat} demonstrate the effectiveness of our approach. For future work, we would like to extend Receiver to conversational recommender systems. After turns of chatting, the agent should be able to infer the user's persona, based on which personalized contents can be recommended.

\section*{Acknowledgments}

We thank all the anonymous reviewers for their valuable comments. This work was supported in part by National Natural Science Foundation of China (U1736217 and 61932003), and National Key R\&D Program of China (2019YFF0302902).

\bibliography{ms}

\begin{thebibliography}{37}
\expandafter\ifx\csname natexlab\endcsname\relax\def\natexlab#1{#1}\fi

\bibitem[{Bahdanau et~al.(2015)Bahdanau, Cho, and
  Bengio}]{Bahdanau2014NeuralMT}
Dzmitry Bahdanau, Kyunghyun Cho, and Yoshua Bengio. 2015.
\newblock \href {http://arxiv.org/abs/1409.0473} {Neural machine translation by
  jointly learning to align and translate}.
\newblock In \emph{Proceedings of the 3rd International Conference on Learning
  Representations, {ICLR} 2015, May 7-9, 2015, San Diego, CA, USA}.

\bibitem[{Devlin et~al.(2019)Devlin, Chang, Lee, and
  Toutanova}]{devlin2018bert}
Jacob Devlin, Ming-Wei Chang, Kenton Lee, and Kristina Toutanova. 2019.
\newblock \href {https://doi.org/10.18653/v1/N19-1423} {{BERT}: Pre-training of
  deep bidirectional transformers for language understanding}.
\newblock In \emph{Proceedings of the 2019 Conference of the North {A}merican
  Chapter of the Association for Computational Linguistics: Human Language
  Technologies, NAACL 2019}, Minneapolis, Minnesota. Association for
  Computational Linguistics.

\bibitem[{Dinan et~al.(2019)Dinan, Logacheva, Malykh, Miller, Shuster, Urbanek,
  Kiela, Szlam, Serban, Lowe, Prabhumoye, Black, Rudnicky, Williams, Pineau,
  Burtsev, and Weston}]{dinan2019second}
Emily Dinan, Varvara Logacheva, Valentin Malykh, Alexander~H. Miller, Kurt
  Shuster, Jack Urbanek, Douwe Kiela, Arthur Szlam, Iulian Serban, Ryan Lowe,
  Shrimai Prabhumoye, Alan~W. Black, Alexander~I. Rudnicky, Jason Williams,
  Joelle Pineau, Mikhail Burtsev, and Jason Weston. 2019.
\newblock \href {http://arxiv.org/abs/1902.00098} {The second conversational
  intelligence challenge (convai2)}.
\newblock \emph{CoRR}, abs/1902.00098.

\bibitem[{Gu et~al.(2019)Gu, Ling, Zhu, and Liu}]{gu2019dually}
Jia-Chen Gu, Zhen-Hua Ling, Xiaodan Zhu, and Quan Liu. 2019.
\newblock \href {https://doi.org/10.18653/v1/D19-1193} {Dually interactive
  matching network for personalized response selection in retrieval-based
  chatbots}.
\newblock In \emph{Proceedings of the 2019 Conference on Empirical Methods in
  Natural Language Processing and the 9th International Joint Conference on
  Natural Language Processing, EMNLP-IJCNLP 2019}, Hong Kong, China.
  Association for Computational Linguistics.

\bibitem[{Hasson et~al.(2012)Hasson, Ghazanfar, Galantucci, Garrod, and
  Keysers}]{hasson2012brain}
Uri Hasson, Asif~A Ghazanfar, Bruno Galantucci, Simon Garrod, and Christian
  Keysers. 2012.
\newblock \href {https://www.ncbi.nlm.nih.gov/pmc/articles/PMC3269540/}
  {Brain-to-brain coupling: a mechanism for creating and sharing a social
  world}.
\newblock \emph{Trends in cognitive sciences}.

\bibitem[{Hinton et~al.(2015)Hinton, Vinyals, and Dean}]{hinton2015distilling}
Geoffrey~E. Hinton, Oriol Vinyals, and Jeffrey Dean. 2015.
\newblock \href {http://arxiv.org/abs/1503.02531} {Distilling the knowledge in
  a neural network}.
\newblock \emph{CoRR}, abs/1503.02531.

\bibitem[{Joshi et~al.(2017)Joshi, Mi, and Faltings}]{joshi2017personalization}
Chaitanya~K. Joshi, Fei Mi, and Boi Faltings. 2017.
\newblock \href {http://arxiv.org/abs/1706.07503} {Personalization in
  goal-oriented dialog}.
\newblock \emph{CoRR}, abs/1706.07503.

\bibitem[{Kingma and Ba(2015)}]{kingma2014adam}
Diederik~P. Kingma and Jimmy Ba. 2015.
\newblock \href {http://arxiv.org/abs/1412.6980} {Adam: {A} method for
  stochastic optimization}.
\newblock In \emph{Proceedings of the 3rd International Conference on Learning
  Representations, {ICLR} 2015, May 7-9, 2015, San Diego, CA, USA}.

\bibitem[{Lewis et~al.(2017)Lewis, Yarats, Dauphin, Parikh, and
  Batra}]{lewis2017deal}
Mike Lewis, Denis Yarats, Yann Dauphin, Devi Parikh, and Dhruv Batra. 2017.
\newblock \href {https://doi.org/10.18653/v1/D17-1259} {Deal or no deal?
  end-to-end learning of negotiation dialogues}.
\newblock In \emph{Proceedings of the 2017 Conference on Empirical Methods in
  Natural Language Processing, EMNLP 2017}, Copenhagen, Denmark. Association
  for Computational Linguistics.

\bibitem[{Li et~al.(2016{\natexlab{a}})Li, Galley, Brockett, Spithourakis, Gao,
  and Dolan}]{li2016persona}
Jiwei Li, Michel Galley, Chris Brockett, Georgios Spithourakis, Jianfeng Gao,
  and Bill Dolan. 2016{\natexlab{a}}.
\newblock \href {https://doi.org/10.18653/v1/P16-1094} {A persona-based neural
  conversation model}.
\newblock In \emph{Proceedings of the 54th Annual Meeting of the Association
  for Computational Linguistics, ACL 2016}, Berlin, Germany. Association for
  Computational Linguistics.

\bibitem[{Li et~al.(2016{\natexlab{b}})Li, Monroe, Ritter, Jurafsky, Galley,
  and Gao}]{li2016deep}
Jiwei Li, Will Monroe, Alan Ritter, Dan Jurafsky, Michel Galley, and Jianfeng
  Gao. 2016{\natexlab{b}}.
\newblock \href {https://doi.org/10.18653/v1/D16-1127} {Deep reinforcement
  learning for dialogue generation}.
\newblock In \emph{Proceedings of the 2016 Conference on Empirical Methods in
  Natural Language Processing, EMNLP 2016}, Austin, Texas. Association for
  Computational Linguistics.

\bibitem[{Liu et~al.(2019)Liu, Chen, Liu, Lou, Fang, Zhou, and
  Zhang}]{liu-etal-2019-split}
Qian Liu, Bei Chen, Haoyan Liu, Jian-Guang Lou, Lei Fang, Bin Zhou, and Dongmei
  Zhang. 2019.
\newblock \href {https://doi.org/10.18653/v1/D19-1535} {A split-and-recombine
  approach for follow-up query analysis}.
\newblock In \emph{Proceedings of the 2019 Conference on Empirical Methods in
  Natural Language Processing and the 9th International Joint Conference on
  Natural Language Processing, EMNLP-IJCNLP 2019}, Hong Kong, China.
  Association for Computational Linguistics.

\bibitem[{Luo et~al.(2019)Luo, Huang, Zeng, Nie, and Sun}]{luo2018learning}
Liangchen Luo, Wenhao Huang, Qi~Zeng, Zaiqing Nie, and Xu~Sun. 2019.
\newblock \href {https://doi.org/10.1609/aaai.v33i01.33016794} {Learning
  personalized end-to-end goal-oriented dialog}.
\newblock In \emph{Proceedings of the 33rd {AAAI} Conference on Artificial
  Intelligence, {AAAI} 2019, The 31st Innovative Applications of Artificial
  Intelligence Conference, {IAAI} 2019, The 9th {AAAI} Symposium on Educational
  Advances in Artificial Intelligence, {EAAI} 2019, January 27 - February 1,
  2019, Honolulu, Hawaii, USA}. {AAAI} Press.

\bibitem[{Madotto et~al.(2019)Madotto, Lin, Wu, and
  Fung}]{madotto2019personalizing}
Andrea Madotto, Zhaojiang Lin, Chien-Sheng Wu, and Pascale Fung. 2019.
\newblock \href {https://doi.org/10.18653/v1/P19-1542} {Personalizing dialogue
  agents via meta-learning}.
\newblock In \emph{Proceedings of the 57th Annual Meeting of the Association
  for Computational Linguistics, ACL 2019}, Florence, Italy. Association for
  Computational Linguistics.

\bibitem[{Mazar{\'e} et~al.(2018)Mazar{\'e}, Humeau, Raison, and
  Bordes}]{mazare2018training}
Pierre-Emmanuel Mazar{\'e}, Samuel Humeau, Martin Raison, and Antoine Bordes.
  2018.
\newblock \href {https://doi.org/10.18653/v1/D18-1298} {Training millions of
  personalized dialogue agents}.
\newblock In \emph{Proceedings of the 2018 Conference on Empirical Methods in
  Natural Language Processing, EMNLP 2018}, Brussels, Belgium. Association for
  Computational Linguistics.

\bibitem[{Miller et~al.(2017)Miller, Feng, Batra, Bordes, Fisch, Lu, Parikh,
  and Weston}]{miller2017parlai}
Alexander Miller, Will Feng, Dhruv Batra, Antoine Bordes, Adam Fisch, Jiasen
  Lu, Devi Parikh, and Jason Weston. 2017.
\newblock \href {https://doi.org/10.18653/v1/D17-2014} {{P}arl{AI}: A dialog
  research software platform}.
\newblock In \emph{Proceedings of the 2017 Conference on Empirical Methods in
  Natural Language Processing: System Demonstrations, EMNLP 2017}, Copenhagen,
  Denmark. Association for Computational Linguistics.

\bibitem[{Olabiyi et~al.(2019)Olabiyi, Khazane, Salimov, and
  Mueller}]{moradversarial}
Oluwatobi Olabiyi, Anish Khazane, Alan Salimov, and Erik~T. Mueller. 2019.
\newblock \href {http://arxiv.org/abs/1905.01992} {An adversarial learning
  framework for a persona-based multi-turn dialogue model}.
\newblock \emph{CoRR}, abs/1905.01992.

\bibitem[{Papineni et~al.(2002)Papineni, Roukos, Ward, and
  Zhu}]{papineni-etal-2002-bleu}
Kishore Papineni, Salim Roukos, Todd Ward, and Wei-Jing Zhu. 2002.
\newblock \href {https://doi.org/10.3115/1073083.1073135} {{BLEU}: a method for
  automatic evaluation of machine translation}.
\newblock In \emph{Proceedings of the 40th Annual Meeting of the Association
  for Computational Linguistics, ACL 2002}, Philadelphia, Pennsylvania, USA.
  Association for Computational Linguistics.

\bibitem[{Paszke et~al.(2019)Paszke, Gross, Massa, Lerer, Bradbury, Chanan,
  Killeen, Lin, Gimelshein, Antiga, Desmaison, Kopf, Yang, DeVito, Raison,
  Tejani, Chilamkurthy, Steiner, Fang, Bai, and Chintala}]{paszke2017automatic}
Adam Paszke, Sam Gross, Francisco Massa, Adam Lerer, James Bradbury, Gregory
  Chanan, Trevor Killeen, Zeming Lin, Natalia Gimelshein, Luca Antiga, Alban
  Desmaison, Andreas Kopf, Edward Yang, Zachary DeVito, Martin Raison, Alykhan
  Tejani, Sasank Chilamkurthy, Benoit Steiner, Lu~Fang, Junjie Bai, and Soumith
  Chintala. 2019.
\newblock \href
  {http://papers.neurips.cc/paper/9015-pytorch-an-imperative-style-high-performance-deep-learning-library.pdf}
  {Pytorch: An imperative style, high-performance deep learning library}.
\newblock In H.~Wallach, H.~Larochelle, A.~Beygelzimer, F.~d\textquotesingle
  Alch\'{e}-Buc, E.~Fox, and R.~Garnett, editors, \emph{Advances in Neural
  Information Processing Systems 32: Annual Conference on Neural Information
  Processing Systems 2019, NeurIPS 2019, December 8-14, 2019, Vancouver, BC,
  Canada}. Curran Associates, Inc.

\bibitem[{Radford et~al.(2018)Radford, Narasimhan, Salimans, and
  Sutskever}]{Radford2018ImprovingLU}
Alec Radford, Karthik Narasimhan, Tim Salimans, and Ilya Sutskever. 2018.
\newblock \href
  {https://s3-us-west-2.amazonaws.com/openai-assets/research-covers/language-unsupervised/language_understanding_paper.pdf}
  {Improving language understanding by generative pre-training}.

\bibitem[{Serban et~al.(2017)Serban, Sordoni, Lowe, Charlin, Pineau, Courville,
  and Bengio}]{serban2017hierarchical}
Iulian~Vlad Serban, Alessandro Sordoni, Ryan Lowe, Laurent Charlin, Joelle
  Pineau, Aaron~C. Courville, and Yoshua Bengio. 2017.
\newblock \href {http://aaai.org/ocs/index.php/AAAI/AAAI17/paper/view/14567} {A
  hierarchical latent variable encoder-decoder model for generating dialogues}.
\newblock In \emph{Proceedings of the 31st {AAAI} Conference on Artificial
  Intelligence, AAAI 2019, February 4-9, 2017, San Francisco, California,
  {USA}}. {AAAI} Press.

\bibitem[{Song et~al.(2020)Song, Zhang, Hu, and Liu}]{song-aaai2020-generating}
Haoyu Song, Wei-Nan Zhang, Jingwen Hu, and Ting Liu. 2020.
\newblock \href {https://arxiv.org/pdf/1911.05889.pdf} {Generating persona
  consistent dialogues by exploiting natural language inference}.
\newblock In \emph{Proceedings of the 34th {AAAI} Conference on Artificial
  Intelligence, AAAI 2020, February 7-12, 2020, New York City, New York, USA}.
  {AAAI} Press.

\bibitem[{Song et~al.(2019)Song, Zhang, Cui, Wang, and Liu}]{song2019diverse}
Haoyu Song, Weinan Zhang, Yiming Cui, Dong Wang, and Ting Liu. 2019.
\newblock \href {https://doi.org/10.24963/ijcai.2019/721} {Exploiting persona
  information for diverse generation of conversational responses}.
\newblock In \emph{Proceedings of the 28th International Joint Conference on
  Artificial Intelligence, {IJCAI} 2019, August 10-16, 2019, Macao, China}.

\bibitem[{Sordoni et~al.(2015)Sordoni, Galley, Auli, Brockett, Ji, Mitchell,
  Nie, Gao, and Dolan}]{sordonineural2015}
Alessandro Sordoni, Michel Galley, Michael Auli, Chris Brockett, Yangfeng Ji,
  Margaret Mitchell, Jian-Yun Nie, Jianfeng Gao, and Bill Dolan. 2015.
\newblock \href {https://doi.org/10.3115/v1/N15-1020} {A neural network
  approach to context-sensitive generation of conversational responses}.
\newblock In \emph{Proceedings of the 2015 Conference of the North {A}merican
  Chapter of the Association for Computational Linguistics: Human Language
  Technologies, NAACL 2015}, Denver, Colorado. Association for Computational
  Linguistics.

\bibitem[{Sutskever et~al.(2014)Sutskever, Vinyals, and
  Le}]{sutskever2014sequence}
Ilya Sutskever, Oriol Vinyals, and Quoc~V. Le. 2014.
\newblock \href
  {http://papers.nips.cc/paper/5346-sequence-to-sequence-learning-with-neural-networks}
  {Sequence to sequence learning with neural networks}.
\newblock In \emph{Advances in Neural Information Processing Systems 27: Annual
  Conference on Neural Information Processing Systems 2014, NIPS 2014, December
  8-13 2014, Montreal, Quebec, Canada}.

\bibitem[{Sutton et~al.(1999)Sutton, McAllester, Singh, and
  Mansour}]{sutton2000policy}
Richard~S. Sutton, David~A. McAllester, Satinder~P. Singh, and Yishay Mansour.
  1999.
\newblock \href
  {http://papers.nips.cc/paper/1713-policy-gradient-methods-for-reinforcement-learning-with-function-approximation}
  {Policy gradient methods for reinforcement learning with function
  approximation}.
\newblock In \emph{Advances in Neural Information Processing Systems 12: Annual
  Conference on Neural Information Processing Systems 1999, NIPS 1999, November
  29 - December 4, 1999, Denver, Colorado, USA,}. The {MIT} Press.

\bibitem[{Tigunova et~al.(2019)Tigunova, Yates, Mirza, and
  Weikum}]{tigunova2019listening}
Anna Tigunova, Andrew Yates, Paramita Mirza, and Gerhard Weikum. 2019.
\newblock \href {https://doi.org/10.1145/3308558.3313498} {Listening between
  the lines: Learning personal attributes from conversations}.
\newblock In \emph{Proceedings of the World Wide Web Conference, {WWW} 2019,
  May 13-17, 2019, San Francisco, CA, USA}. {ACM}.

\bibitem[{Vaswani et~al.(2017)Vaswani, Shazeer, Parmar, Uszkoreit, Jones,
  Gomez, Kaiser, and Polosukhin}]{vaswani2017attention}
Ashish Vaswani, Noam Shazeer, Niki Parmar, Jakob Uszkoreit, Llion Jones,
  Aidan~N. Gomez, Lukasz Kaiser, and Illia Polosukhin. 2017.
\newblock \href {http://papers.nips.cc/paper/7181-attention-is-all-you-need}
  {Attention is all you need}.
\newblock In \emph{Advances in Neural Information Processing Systems 30: Annual
  Conference on Neural Information Processing Systems 2017, NIPS 2017, December
  4-9, 2017, Long Beach, CA, {USA}}.

\bibitem[{Weaver and Tao(2001)}]{weaver2001optimal}
Lex Weaver and Nigel Tao. 2001.
\newblock \href
  {https://dslpitt.org/uai/displayArticleDetails.jsp?mmnu=1\&smnu=2\&article\_id=141\&proceeding\_id=17}
  {The optimal reward baseline for gradient-based reinforcement learning}.
\newblock In \emph{Proceedings of the 17th Conference in Uncertainty in
  Artificial Intelligence, {UAI} 2001, August 2-5, 2001, University of
  Washington, Seattle, Washington, USA}. Morgan Kaufmann.

\bibitem[{Williams(1992)}]{williams1992simple}
Ronald~J. Williams. 1992.
\newblock \href {https://doi.org/10.1007/BF00992696} {Simple statistical
  gradient-following algorithms for connectionist reinforcement learning}.
\newblock \emph{Machine learning}.

\bibitem[{Wolf et~al.(2019{\natexlab{a}})Wolf, Debut, Sanh, Chaumond, Delangue,
  Moi, Cistac, Rault, Louf, Funtowicz, and Brew}]{Wolf2019HuggingFacesTS}
Thomas Wolf, Lysandre Debut, Victor Sanh, Julien Chaumond, Clement Delangue,
  Anthony Moi, Pierric Cistac, Tim Rault, R{\'{e}}mi Louf, Morgan Funtowicz,
  and Jamie Brew. 2019{\natexlab{a}}.
\newblock \href {http://arxiv.org/abs/1910.03771} {Huggingface's transformers:
  State-of-the-art natural language processing}.
\newblock \emph{CoRR}, abs/1910.03771.

\bibitem[{Wolf et~al.(2019{\natexlab{b}})Wolf, Sanh, Chaumond, and
  Delangue}]{wolf2019transfertransfo}
Thomas Wolf, Victor Sanh, Julien Chaumond, and Clement Delangue.
  2019{\natexlab{b}}.
\newblock \href {http://arxiv.org/abs/1901.08149} {{TransferTransfo}: A
  transfer learning approach for neural network based conversational agents}.
\newblock \emph{CoRR}, abs/1901.08149.

\bibitem[{Wu et~al.(2017)Wu, Wu, Xing, Zhou, and Li}]{wu2017sequential}
Yu~Wu, Wei Wu, Chen Xing, Ming Zhou, and Zhoujun Li. 2017.
\newblock \href {https://doi.org/10.18653/v1/P17-1046} {Sequential matching
  network: A new architecture for multi-turn response selection in
  retrieval-based chatbots}.
\newblock In \emph{Proceedings of the 55th Annual Meeting of the Association
  for Computational Linguistics, ACL 2017}, Vancouver, Canada. Association for
  Computational Linguistics.

\bibitem[{Yang et~al.(2018)Yang, Tu, Qu, Zhao, Chen, and
  Zhu}]{yang2018personalized}
Min Yang, Wenting Tu, Qiang Qu, Zhou Zhao, Xiaojun Chen, and Jia Zhu. 2018.
\newblock \href {https://doi.org/10.1016/j.neunet.2018.03.009} {Personalized
  response generation by dual-learning based domain adaptation}.
\newblock \emph{Neural Networks}.

\bibitem[{Zemlyanskiy and Sha(2018)}]{zemlyanskiy2018aiming}
Yury Zemlyanskiy and Fei Sha. 2018.
\newblock \href {https://doi.org/10.18653/v1/K18-1053} {Aiming to know you
  better perhaps makes me a more engaging dialogue partner}.
\newblock In \emph{Proceedings of the 22nd Conference on Computational Natural
  Language Learning, CoNLL 2018}, Brussels, Belgium. Association for
  Computational Linguistics.

\bibitem[{Zhang et~al.(2018{\natexlab{a}})Zhang, Lan, Guo, Xu, and
  Cheng}]{zhang2018reinforcing}
Hainan Zhang, Yanyan Lan, Jiafeng Guo, Jun Xu, and Xueqi Cheng.
  2018{\natexlab{a}}.
\newblock \href {https://doi.org/10.24963/ijcai.2018/635} {Reinforcing
  coherence for sequence to sequence model in dialogue generation}.
\newblock In \emph{Proceedings of the 27th International Joint Conference on
  Artificial Intelligence, {IJCAI} 2018, July 13-19, 2018, Stockholm, Sweden}.

\bibitem[{Zhang et~al.(2018{\natexlab{b}})Zhang, Dinan, Urbanek, Szlam, Kiela,
  and Weston}]{zhang2018personalizing}
Saizheng Zhang, Emily Dinan, Jack Urbanek, Arthur Szlam, Douwe Kiela, and Jason
  Weston. 2018{\natexlab{b}}.
\newblock \href {https://doi.org/10.18653/v1/P18-1205} {Personalizing dialogue
  agents: {I} have a dog, do you have pets too?}
\newblock In \emph{Proceedings of the 56th Annual Meeting of the Association
  for Computational Linguistics, ACL 2018}, Melbourne, Australia. Association
  for Computational Linguistics.

\end{thebibliography}
\bibliographystyle{acl_natbib}

\end{document}